\documentclass[11pt,a4paper]{article}
\usepackage[hyperref]{emnlp2018}
\usepackage{times}
\usepackage{latexsym}
\usepackage{float}
\restylefloat{table}
\usepackage{xcolor}
\usepackage{graphicx}
\usepackage{adjustbox}
\usepackage{makecell}
\usepackage{tabularx}

\usepackage{url}

\aclfinalcopy 

\title{A Reinforcement Learning-driven Translation Model for Search-Oriented Conversational Systems}

  \author{Wafa Aissa \\
 Sorbonne Universit\'e\\
 CNRS, LIP6\\
F-75005 Paris, France\\
  {\small \tt wafa.aissa@lip6.fr} \\\And
  Laure Soulier \\
  Sorbonne Universit\'e\\
  CNRS, LIP6\\
F-75005 Paris, France\\
  {\small \tt laure.soulier@lip6.fr} \\ \And
   Ludovic Denoyer \\
  Sorbonne Universit\'e\\
  CNRS, LIP6\\
F-75005 Paris, France\\
  {\small \tt ludovic.denoyer@lip6.fr}
  }

\date{}

\begin{document}
\maketitle

\begin{abstract}
 Search-oriented conversational systems rely on information needs expressed in natural language (NL). We focus here on the understanding of NL expressions for building keyword-based queries. We propose a  reinforcement-learning-driven translation model framework  able to 1) learn the translation from NL expressions to queries in a supervised way, and, 2) to overcome the lack of large-scale dataset by framing the translation model as a word selection approach and injecting relevance feedback in the learning process. Experiments are carried out on two TREC datasets and outline the effectiveness of our approach.
\end{abstract}

\section{Introduction}

Artificial Intelligence, and more particularly deep learning,  have recently opened tremendous perspectives for reasoning over semantics in text-based applications such as machine translation \cite{Lample2017}, chat-bot \cite{Bordes2016}, knowledge base completion \cite{Lin:2015} or extraction \cite{Hoffmann2011}. 
Very recently, conversational information retrieval (IR) has emerged as a new paradigm in IR \cite{Burtsev2017SCAI,Joho2018}, in which natural conversations between humans and computers are used to  satisfy an information need. As for now, conversational systems are limited to simple conversational interactions (namely, chit-chat conversations) \cite{Jiwei2016,Ritter2011}, closed worlds driven by domain-adapted  or slot-filling patterns \cite{Bordes2016,Wang2013} (e.g., a travel planning task requiring to book a flight, then a hotel, etc...), or knowledge-base extraction (e.g., information extraction tasks) \cite{Dhingra2017}.

In contrast, search-oriented conversational systems (SOCS) aim at finding information in an open world (both unstructured information sources and knowledge-bases) in response to users' information needs expressed in natural language (NL); the latter often being ambiguous. Therefore, one key challenge of SOCS is to understand  users' information needs expressed in NL to identify relevant documents. 

Formulating  an information need through queries has been outlined as a difficult task \cite{Vakulenko2017,Agichtein2006,Joachims2002} which is generally tackled by refining/reformulating queries using pseudo-relevance feedback or users' clicks. In SOCS, there is an upstream challenge dealing with the building of the query from a NL expression that initiates the search session to avoid useless users' interactions with the system.  
This problem could be tackled for instance through deep neural translation models (e.g., encoder-decoder approaches) 
as initiated by \cite{Song2017,Yin2017}. 
However, these methods learn the query formulation model independently of the search task at hand. To overpass this limitation, \cite{Nogueira2017} have proposed a reinforcement learning model for query reformulation in which the reward is based on  terms of documents retrieved by the IR system. \\ 
\begin{figure*}
    \centering
    \includegraphics[scale=0.4]{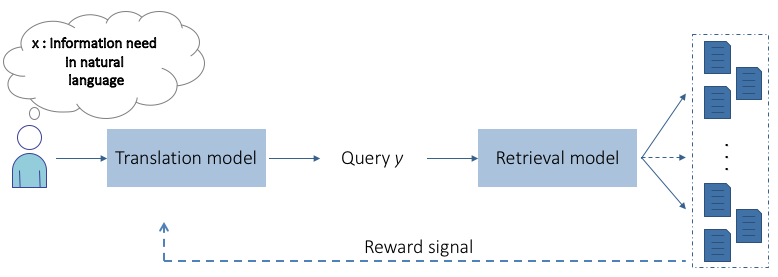}
    \caption{Overview of our reinforcement learning-driven translation model for SOCS}
    \label{fig:overview}
\end{figure*}

In this work, we propose to bridge these two lines of work: 1) machine translation to learn the mapping between information needs expressed in NL and information needs  formulated using keywords \cite{Song2017,Yin2017}, and 2) reinforcement learning to inject the task objectives within the machine translation model \cite{Nogueira2017}. 
More particularly, we propose a two-step model which first learns the translation model through the supervision of  NL-query pairs and then refines the translation model using a relevance feedback provided by the search engine. It is worth mentioning that there does not exist SOCS-oriented dataset that both aligns users' information needs in NL with keyword-based queries and includes a document collection to perform a retrieval task. To the best of our knowledge, TREC datasets are the only ones expressing such constraint, but the number of NL-query pairs is however limited. To fit with the issue of dealing with large vocabulary and the dataset constraint, we frame the translation model as a word selection one which aims at identifying which words in the NL expression can be used to build the query. 
Our model is evaluated on two TREC datasets. The obtained results outline the effectiveness of combining reinforcement learning  with machine translation models. \\

The remaining of the paper is organized as follows. Section 2 details our translation model. Section 3 presents the evaluation protocol and results are highlighted in Section 4. The conclusion and perspectives are discussed in Section 5.

\section{Reinforcement learning-driven translation model}

\subsection{Notation and problem formulation}
Our reinforcement learning-driven translation model allows to formulate a user's information need $x$ expressed in NL into a keyword-based query $y$. The user's information need $x$ is a sequence of $n$ words ($x=x_1, ..., x_i, ..., x_n$). 
To fit with our word selection objective, the  query $y$ is modeled as a binary vector $y \in \{0,1\}^{n}$ of size $n$ (namely, the size of the  natural language expression $x$). Each element $y_j \in y$ equals to 1 if word $x_i \in x$ exists in query $y$ and 0 otherwise. For example, if we consider the NL as "Identify documents that discuss sick building syndrome or  building related illnesses." and the key-words query as "sick building syndrome.", the expected query will be formulated as follows:    $y=(0,0,0,0,1,1,1,0,0,0,0)$ .  

The objective of our model $f_{\theta}$ (with $\theta$ being the parameters of our model) is  to estimate the probability  $p(y|x)$ of generating the binary vector $y$ given the NL expression $x$. Since terms are not independent within the formulation of NL expressions and queries, it makes sense to consider that the selection of a word  is conditioned by the sequence of decisions taken on previous words $y_{<i}$. Thus, $P(y|x)$ could be written as follows:
\begin{equation}
    p(y|x)=  \prod_{y_i \in y} p(y_i|y_{<i}, x)
\end{equation}

This probability is first learned using a maximum likelihood estimation (MLE) on the basis of NL-query pairs (Section 2.2). Then, this probability is refined using reinforcement learning techniques (Section 2.3). We end up with the network architecture used in the translation model.

\begin{figure*}
    \centering
    \includegraphics[scale=0.5]{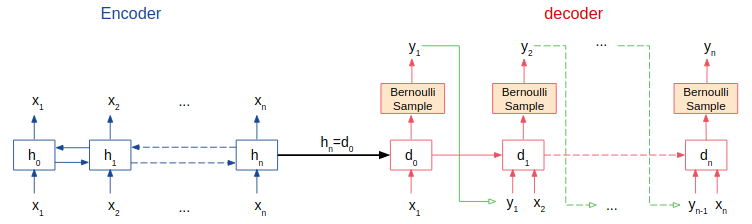}
    \caption{Network architecture of our translation model}
    \label{fig:archi}
\end{figure*}

\subsection{Supervised translation model: from NL to queries}
The translation model works as a supervised word selection model aiming at building queries $y$ by using the vocabulary available in NL expressions $x$. 
To do so, we use a set $D$ of $N$ NL-query pairs $D = \{(x^{1}, y^{1}), ..., (x^{k},y^{k}), ...,  (x^{N},y^{N})\}$.

The objective of the translation model is to predict whether each word $x_i^{k}$ in the NL expression $x^{k}$ is included in the expected query $y^{k}$. In other words, it consists in predicting the probability $p(\hat{y}_i^{k}=y_i^{k} | \hat{y}_{<i}^{k}, x^{k})$ that the $i^{th}$ element $\hat{y}_i^{k}$ of vector $\hat{y}^{k}$ is equal to the same element $y_i^{k}$ in the original query $y^{k}$ (namely, that $\hat{y}_i^{k}=y_i^{k}$) given the state of previous elements $\hat{y}_{<i}^{k}$ and the NL expression $x^{k}$. This  probability $p(\hat{y}_i^{k}=y_i^{k} | \hat{y}_{<i}^{k}, x^{k})$ is modeled using a Bernoulli distribution in which parameters  are estimated through the probability distribution. 

Let's define for a NL-query instance $(x^{k},y^{k})$,  $f_{(\theta,x^{k})} =  \sum_{y_{i}^{k} \in y^{k}} log(p(\hat{y}_i^{k}=y_i^{k} | \hat{y}_{<i}^{k}, x^{k}))$. 
The translation model is trained by maximizing the following MLE over the set $D$ of  NL-query pairs $(x_k,y_k)$:  
\begin{equation}
L_{SMT} = \sum_{(x^{k},y^{k}) \in D} log (f(\theta,x^{k}))
\end{equation}

\subsection{Reinforcement learning}
To inject the task objective in the translation model, we consider that the process of query building could be enhanced through reinforcement learning techniques. Therefore, the word selection could be seen as a sequence of choices of selecting word $x_t$ at each time step $t$. 
The choices are \textit{rewarded} at the end of the selection process by a metric measuring the effectiveness of the query building process within a retrieval task.  Particularly, the predicted query $\hat{y}$ obtained from the binary vector $\hat{y}$  is fed to a retrieval model to rank documents. For each NL expression $x$ (and accordingly the associated predicted  query $\hat{y}$), we dispose of a set $\mathcal{D}_x$ of relevant documents  (also called ground truth). We note $GT$ the set of $n$ pairs $(x;\mathcal{D}_x)$. With this in mind, the effectiveness of the obtained ranking could be estimated using an effectiveness-driven metric (e.g., the MAP). Thus, the reward $R$ for a generated query $\hat{y}$ given the relevance feedback pair $(x,\mathcal{D}_x)$ is obtained as follows: 
\begin{equation}
    R(\hat{y}) = MAP(\hat{y},\mathcal{D}_x)
\end{equation}

At the end of the selection process, the objective function aims at maximizing the expectation of the search effectiveness over the predicted queries:
\begin{equation}
    L_{RL}(\theta)= arg\max_{\theta}~  \mathrm{E}_{\stackrel{(x;\mathcal{D}_x) \in GT}{ \hat{y} \sim  f_{\theta}(x)}} ~ [R(\hat{y})]
\end{equation}
where $\hat{y}$ is given by the translation model $f_{\theta}(x)$.
This objective function is maximized  using gradient descent techniques \cite{Baxter1999}.

\subsection{Model architecture}

    The model is based on an encoder-decoder  building a query $\hat{q}$ from the input $x$. 
    Particularly, each element $x_i$ of $x$ is modeled through word embeddings $w_{x_i}$; resulting in a sequence $w_x$ of word embeddings for input $x$.
    As shown in Figure \ref{fig:archi}, the encoder is a bi-directional LSTM \cite{Hochreiter:1997:LSM:1246443.1246450}  aiming to transform the input sequence $w_x$ to its continuous representation $h_n$.
    The decoder is composed of a LSTM  in which each word $x_i$ is injected to estimate the word selection probability $p(y_i|y_{<i}, x)$ using the hidden vector $h_n$ learned in the encoder network and the current word $x_i$; leading to estimate probability  $p(y_i|y_{<i},x_i,h_n)$.

\begin{table*}[t]
\centering
\small
\begin{tabular}{|l|l|c|c|c|c|}
\hline
 {\bf TREC track} & {\bf collection} & {\bf  pairs} & {\bf NL length} & {\bf avg of duplic. word  in NL} \\\hline
 TREC Robust (2004) & disk4-5 & 250 & 15.333 & 1.108 \\ 
TREC Web (2000 2001) & WT10G & 100 & 11.47 & 0.65 \\\hline
\end{tabular}
\caption{Dataset statistics separated per document collections}
\label{tab:stat}
\end{table*}

\section{Protocol design}

\subsection{Datasets}
Since there does not exist yet SOCS-driven datasets including NL-query pairs, we use TREC  tracks (namely, Robust 2004 and Web 2000-2001). In these tracks, query topics include a title, a topic description and a narrative text; the two latter being formulated in natural language.  To build query-NL pairs, we  use the title to form the set of keyword queries and the description for the set of information needs expressed in NL. An example of a query-NL pair is:
{\small
   \begin{tabular}{|l|p{5cm}|}
  \hline
  Title & Lewis and Clark expedition \\
  \hline
  Description & What are some useful sites containing information about the historic Lewis and Clark expedition? \\
  \hline
\end{tabular}}

This NL-query building process results in 350 pairs in total as presented in Table \ref{tab:stat}. 

We are aware that the use of TREC datasets is biased in the sense that it does not exactly fit with the expression of NL information need in the context of conversational systems, but we believe that the description is enough verbose to evaluate the impact of our query building model in this exploratory work. Further experiments with generated datasets, as done in \cite{Song2017}, will be carried out in the future.

We also analyze the issue of duplicate words into TREC descriptions since it can directly impact the query formulation process based on word selection in the  word sequence of TREC descriptions. In practice, this might lead to select several times the same word to build the query, and, therefore, directly impact the retrieval performance.  As shown in Table \ref{tab:stat}, the ratio of duplicate words in TREC descriptions over the whole set of queries is very low (1.1 duplicate words in average in each query for TREC Robust and 0.65 for TREC Web). This  suggests that this issue is minor in the used datasets. We, therefore, decided to skip this issue for the moment.

\subsection{Metrics and baselines}
To evaluate our approach, we measure the retrieval effectiveness of the predicted queries. To do so, for each predicted query, we run the BM25 model through an IR system (namely, PyLucene\footnote{\url{http://lucene.apache.org/pylucene/}}) to obtain a document ranking. The latter is evaluated through the MAP metric.\\ 

To show the soundness of our approach (namely, translating information needs expressed in NL into queries), we compare our generated queries to scenario \textbf{NL} feeding  the natural language information needs (TREC descriptions in our protocol) to the IR retrieval system.

Since the objective of our model is to formulate queries, we also evaluate the effectiveness of original TREC titles (scenario \textbf{Q}). This setting rather refers to the oracle that our model must reach. 

We  mentioned that before training the selection model we transformed each $x$ to its binary representation $y$ based on the presence of the words in the ground truth query. The dataset being slightly biased by this binary modeling, we observed that not all the words existing in the query  do exist in $x$. To analyze this bias, we also compare our approach with these binary queries (scenario \textbf{Q bin}) referring to the projection of queries \textbf{Q} on the vocabulary available in the \textbf{NL} description.
 
We also compare our model to a random approach which randomly selects 3 words from $x$ to build queries (scenario \textbf{Random}).\\

Different variants of our model are also tested:
\begin{itemize}
    \item \textbf{SMT} which only considers the first component of our model based on a supervised machine translation approach (Section 2.2). This variant could be assimilated to the approach proposed in \cite{Song2017} in the sense that the machine translation is performed independently of the task objective.
    \item \textbf{RL} which only considers the reinforcement learning objective function (Section 2.3) without pre-training of the supervised translation model.
    \item \textbf{SMT+RL} which is our full model in which we start by pre-training the model using the supervised translation model (Section 2.3), and, then, we inject the reward signal in the translation probabilities (Section 2.4).
\end{itemize}

\begin{table*}[t]
\centering

\small
\begin{tabular}{|l|c|l|c|l|} 
   \hline
    Baseline &  \multicolumn{2}{c|}{TREC Robust(2004)} & \multicolumn{2}{c|}{ TREC Web (2000-2001)}  \\
    \cline{2-5} 
        & MAP & $ \% Chg $ & MAP & $ \% Chg $  \\
    \hline
    NL &  0.08925 & +15.25\%  *** & 0.15913 & +12.88\%  * \\
    \hline
    Q & 0.09804 & +4.92\%    & 0.16543 & +8.58\%  \\
    \hline
    Q bin & 0.08847 & +16.26\%  * & 0.17402 & +3.22\%   \\
    \hline
    Random & 0.01808 & +468.91\%  *** & 0.04060 & +342.44\%  ***\\
    \hline
    SMT & 0.06845 & +50.27\%  *** & 0.08891 & +102.04\%  *** \\
    \hline
    RL & 0.08983 & +14.51\%  *** & 0.16474 & +9.04\%   \\
    \hline
    \textbf{SMT+RL} & \textbf{0.10286} &    & \textbf{0.17963} &   \\
    \hline
\end{tabular}
\caption{Comparative effectiveness analysis of our approach. $\% Chg $: improvement of \textbf{SMT+RL} over corresponding baselines. Paired t-test significance *: $0.01 < t \leq 0.05$ ; **: $0.001 < t \leq 0.01$ ; ***: $t \leq 0.001$.}
\label{tab:res}
\end{table*}

\begin{table*}[h]
\small
\centering
   \begin{tabular}{|p{5cm}|p{3cm}|p{3cm}|p{3cm}|}
   \hline \textbf{NL} & \textbf{Q} & \textbf{Q bin} & \textbf{SMT+RL} \\\hline 
    what are new methods of producing steel & steel producing & producing steel & new methods of producing steel \\ \hline
        
       what are the advantages and or disadvantages of tooth implant & implant dentistry & implant & advantages disadvantages tooth implant \\ \hline 
        
        find documents that discuss the toronto film festival awards & toronto film awards & toronto film awards & the toronto film festival awards \\ \hline 
        
        find documents that give growth rates of pine trees & where can i find growth rates for the pine trees & growth rates pine trees  &  growth rates of pine trees \\ \hline
        
   \end{tabular}
   \caption{Examples of query formulation for \textbf{NL} queries, the original query \textbf{Q}, the binary version \textbf{Q bin} of the original query, and our model \textbf{SMT-RL}.}
   \label{tab:exmpl}
\end{table*}

\subsection{Implementation details}

To transform each word $x_i$ to its vector representation $w_{x_i}$, we use Fasttext \footnote{\url{https://github.com/facebookresearch/fastText/}} \cite{bojanowski2017enriching} pre-trained word embeddings. 
The encoder and decoders have one hidden layer with 100 hidden units each. 

To train our model, we perform 10-fold cross-validation. 
For the \textbf{SMT+RL} model, we start by a pre-training using the supervised translation model for 100 iterations. The training is then pursued by 1000 iterations while including the reinforcement learning approach. In the latter, the reward, namely the MAP metric, is estimated over document rankings obtained by the BM25 model in PyLucene. We use a minibatch Adam \cite{Kingma2014AdamAM} algorithm to pre-train the model and SGD for the reinforcement learning part. Each update is computed after a minibatch of 12 sentences. 

\section{Results}
We present here the effectiveness of our approach aiming at generating queries from users' information needs expressed in NL.
In Table \ref{tab:res}, we present the  retrieval effectiveness (regarding the MAP) of our model  and the different baselines (\textbf{NL}, \textbf{Q}, \textbf{Q bin}, \textbf{Random}, \textbf{SMT}, and \textbf{RL}) described in section 3.2. 
From a general point of view, results highlight that in both datasets, our proposed model \textbf{SMT+RL} outperforms the different baselines with improvements that are generally significant, ranging from $+3.22\%$ to $+468.91\%$. \\

More particularly, the effectiveness analysis allows to 
draw the following statements:

$\bullet$  The overall performance of the compared approaches generally outperforms the retrieval effectiveness of the \textbf{NL} baseline.  For instance, on TREC Robust,  queries generated by our model allows to  significantly improve the retrieval performance of $+15.25\%$ regarding information needs expressed in NL (MAP: 0.10286 vs. 0.08925). This result validates the motivation of this work to formulate queries from NL expressions. This is relatively intuitive since NL expressions are verbose by nature and might include non-specific words willing to inject noise in the retrieval process. 

$\bullet$  Our approach \textbf{SMT+RL} provides similar results as the \textbf{Q} and \textbf{Q bin}.  Since the objective function of our model is guided by the initial query \textbf{Q} transformed in a binary vector (\textbf{Q bin}), these baselines could be considered as oracles. We note however that our model obtains higher results (improvements from $+3.22\%$ to $+16.26\%$) with a significant difference for the \textbf{Q bin} baseline for TREC Robust. To get a better understanding to what extent our generated queries are different from those used in baselines \textbf{Q} and \textbf{Q bin}, we illustrate in Table \ref{tab:exmpl} some examples. While queries in \textbf{Q}  identify the most important words leading to an exploratory query (e.g. ``steel productions''), our model \textbf{SMT+RL} provides additional words that precise which facet of the query is concerned (e.g., ``new methods of...''), and accordingly  improves the ranking of documents.

$\bullet$ Our model \textbf{SMT+RL} is significantly higher than the \textbf{SMT} baseline which converges to a relatively low MAP value (0.06845 and 0.08891 for TREC Robust and TREC Web, respectively). This could be explained by the fact that our datasets are very small (250 and 100 NL-query pairs respectively for TREC Robust and TREC Web) and that such machine translation approaches are well-known to be data hungry. Reinforcement learning techniques could be a solution to overpass this problem since they inject additional information (namely, the reward) in the network learning.

$\bullet$ The \textbf{RL} baseline achieves relatively good retrieval performances. As we can see from TREC Web, the \textbf{RL} model obtains a MAP of 0.16474 against 0.15913 for the \textbf{NL} baseline. 
The \textbf{RL} baseline allows approaching the retrieval performances of baselines \textbf{Q}  and \textbf{Q bin}, although it obtains lower results. This reinforces our intuition that 1) applying machine translation approaches should be driven by the task (retrieval task in our context) and 2) reinforcement learning techniques provide good strategies to build effective queries. The latter statement has also been outlined in previous work \cite{Nogueira2017}.\\

$\bullet$ The comparison of our model \textbf{SMT+RL} regarding \textbf{SMT} and \textbf{RL} baselines outlines that reinforcement learning techniques might be more beneficial when a pre-training is performed. In our context, the pre-training is performed using the \textbf{SMT} model (Section 2.3) which helps the model to be more general and effective before using the reward signal to guide the selection process.

It is worth mentioning that we also trained in preliminary experiments a state of the art translation models such as a generative encoder-decoder RNN with attention mechanism, as done in \cite{Yin2017, Song2017}. We did not report the results since the model was not able to generalize in the testing phase  over new samples from the  NL-query dataset used in the training phase. This is probably due to the trade-off between the number of training pairs and  the large size of the vocabulary which is not enough represented in different contexts. However, we believe that combining reinforcement learning with attention-mechanism for query-generation is promising. We let this perspective for future work.

\section{Conclusion and future work}
We propose a selection model to transform  user's need in NL into a keyword query to increase the retrieval effectiveness in a SOCS context. Our model bridges two lines of work dealing with supervised machine translation and reinforcement learning. 
Our model has been evaluated using two different TREC datasets and outlines promising results in terms of effectiveness. Our approach has some limitations we plan to overcome in the future. First, our model is framed as a word selection process that could be turned into a generative model. Second, experiments are carried out on small datasets (250 and 100 NL-query pairs) that could be augmented using the evaluation protocol proposed in \cite{Song2017}. In long term, we plan to adapt our model by totally skipping the query formulation step and designing retrieval models dealing with NL expressions.

\bibliographystyle{acl_natbib_nourl}
\bibliography{emnlp2018}

\end{document}